\documentclass[times, review, 10pt]{elsarticle}

\usepackage{amssymb}
\usepackage{amsmath}
\usepackage[numbers]{natbib}
\usepackage{hyperref}
\usepackage{float}
\usepackage{verbatim} 
\restylefloat{figure}
\floatstyle{plaintop} 
\restylefloat{table}
\usepackage{multirow}
\usepackage{tabularx}
\usepackage{adjustbox}
\usepackage{amsmath}
\usepackage{booktabs}
\usepackage{bm}
\usepackage{amssymb}
\usepackage{tabularx}
\usepackage{makecell}
\usepackage{booktabs}
\usepackage{graphicx}
\usepackage{subcaption}
\usepackage{algorithm}
\usepackage{algpseudocode}
\usepackage{siunitx}

\journal{Pattern Recognition}

\begin{document}

\begin{frontmatter}



\title{CMIS-Net: A Cascaded Multi-Scale Individual Standardization Network for Backchannel Agreement Estimation} 

\author[1,2]{Yuxuan Huang} 

\affiliation[1]{organization={Key Laboratory of Advanced Medical Imaging and Intelligent Computing of Guizhou Province, Engineering Research Center of Text Computing, Ministry of Education, State Key Laboratory of Public Big Data, College of Computer Science and Technology, Guizhou University},
            city={Guiyang},
            postcode={550025}, 
            country={China}}
\affiliation[2]{
organization={Department of Rehabilitation Sciences, The Hong Kong Polytechnic University},
city={Hong Kong},
country={China}
}

\author[3]{Kangzhong Wang}
\affiliation[3]{
organization={Department of Computing, The Hong Kong Polytechnic University},
city={Hong Kong},
country={China}
}

\author[4]
{Eugene Yujun Fu \corref{cor1}}
\ead{eugenefu@eduhk.hk}
\affiliation[4]{
organization={Centre for Learning, Teaching and Technology, The Education University of Hong Kong},
city={Hong Kong},
country={China}
}


\author[3]{\\Grace Ngai}

\author[2,3]{Peter H.F. Ng}

\cortext[cor1]{Corresponding author}

\begin{abstract}

Backchannels are subtle listener responses, such as nods, smiles, or short verbal cues like “yes” or “uh-huh,” which convey understanding and agreement in conversations. These signals provide feedback to speakers, improve the smoothness of interaction, and play a crucial role in developing human-like, responsive AI systems. However, the expression of backchannel behaviors is often significantly influenced by individual differences, operating across multiple scales: from instant dynamics such as response intensity (frame-level) to temporal patterns such as frequency and rhythm preferences (sequence-level). This presents a complex pattern recognition problem that contemporary emotion recognition methods have yet to fully address. Particularly, existing individualized methods in emotion recognition often operate at a single scale, overlooking the complementary nature of multi-scale behavioral cues. To address these challenges, we propose a novel Cascaded Multi-Scale Individual Standardization Network (CMIS-Net) that extracts individual-normalized backchannel features by removing person-specific neutral baselines from observed expressions. Operating at both frame and sequence levels, this normalization allows model to focus on relative changes from each person’s baseline rather than absolute expression values. Furthermore, we introduce an implicit data augmentation module to address the observed training data distributional bias, improving model generalization. Comprehensive experiments and visualizations demonstrate that CMIS-Net effectively handles individual differences and data imbalance, achieving state-of-the-art performance in backchannel agreement detection. 
Nevertheless, we believe that the effectiveness of individual standardization for subtle sequential pattern recognition holds the potential to offer a generalizable approach for various pattern recognition tasks that involve person-specific variations.
Our code can be found in: https://github.com/AffectiveComputingLab-HK/CMIS-Net.
\end{abstract}




\end{frontmatter}



\section{Introduction}
\label{sec1}
Recognizing subtle and short-lived patterns in sequential data is a general task in pattern recognition\cite{koller2019weakly}. Among these subtle sequential patterns, backchannel cues serve as a representative example in conversational interactions. Backchannel refer to listener’s non-verbal signals and brief verbal responses during a conversation, such as nodding and smiling or uttering short acknowledgments like "yes" or "uh-huh". As a common communication strategy, backchannels enhance the fluency and effectiveness of interactions, facilitating mutual understanding and reducing potential communication breakdowns \cite{kendon1967some}. 
In addition to maintaining conversational flow, backchannel behaviors reveal the listener's attitude -- agreeing or disagreeing with the speaker's statements. This subjective response, quantified as ''Backchannel Agreement'', provides crucial insights into participants' opinions and interpersonal dynamics within conversations or group meetings \cite{bevacqua2010multimodal,muller2022multimediate}. In face-to-face interactions, listeners' backchannel responses significantly influence speakers' verbal behavior, enabling speakers to sense listeners' engagement and agreement levels and adjust their communication strategies accordingly. 

Investigating backchannel behaviors and agreement patterns holds significant practical value across diverse domains.
For example, in clinical contexts, psychotherapists who observe and guide patients' backchannel responses gain deeper insights into their mental states, enabling more effective treatment strategies and improved therapeutic outcomes \cite{sacks1974simplest}. 
Beyond clinical settings, backchannel analysis enhances our understanding of nonverbal communication in interpersonal interactions. This will result in benefits including more effective communication strategies, enhanced negotiation and conflict resolution \cite{demirdjian20023, goldberg2005secrets}, improved classroom performance \cite{fuchs1987examiner}, and higher quality of childcare interactions \cite{burns1984rapport}.

Moreover, with the rapid advancement of artificial intelligence, enabling machines to accurately understand users' conversational engagement and attitudes has become increasingly critical for developing responsive dialogue systems and robots. The integration of backchannel agreement detection into these systems and robots represents a significant step toward more natural human-machine interactions. While linguistic and social psychological research has long recognized the significance of backchannel behavior analysis, traditional detection methods remain constrained by their reliance on expert knowledge and manual annotation. In contrast, automatic backchannel detection offers compelling advantages in efficiency, real-time responsiveness, and scalability, making it as a promising direction for practical AI applications.

Previous studies \cite{muller2022multimediate, sharma2022graph, amer2023backchannel, wang2023unveiling} have explored various deep learning architectures particularly with visual features (e.g., facial features) for automatic backchannel detection and aggreement estimation. 
However, these approaches face a fundamental challenge: individuals exhibit distinct facial features and behavioral patterns in neutral states and backchannel expressions. These person-specific characteristics can interfere with model predictions, leading to classification and estimation errors.
Nevertheless, existing research has largely overlooked the modeling of individual differences in backchannel agreement detection.

Furthermore, sequential pattern recognition often operates across multiple temporal scales~\cite{liao2024sequence, li2023multi}, particularly from instantaneous response intensity at the frame level to frequency and rhythmic patterns at the sequence level. Both of them convey crucial information about listener's agreement level. However, current individual standardization methods \cite{fan2022isnet} typically operate at a single scale, failing to leverage the complementary nature of multi-scale behavioral cues.
This single-scale limitation significantly constrains model performance. While such approaches have achieved success in various emotion recognition tasks \cite{fan2022isnet,chen2024dsnet}, they inherently suffer from restricted representational capacity: frame-level models may capture fine-grained local details but miss broader contextual patterns, while sequence-level models may emphasize global trends at the expense of subtle momentary cues. This trade-off becomes particularly problematic in backchannel agreement estimation, where both micro-expressions occured in a frame and discourse-level patterns spanning several frames and seconds contribute essential information about the listener's agreement levels.

Moreover, some studies have found that backchannel responses predominantly cluster around neutral to mildly positive expressions, often manifesting as subtle politeness markers like gentle smiling, while strongly emotional backchannels are much less common \cite{reece2023candor}.
This pattern is reinforced by social convention contexts, where individuals typically express agreement through polite and understated responses rather than intense reactions\cite{brown1987politeness, cutrone2005case}.
This tendency leads to an imbalanced distribution of existing backchannel agreement datasets \cite{muller2022multimediate}, with most samples concentrated in the neutral-to-mild range. 
Consequently, models may overfit to the high-density samples of the distribution, reducing their ability to generalize to emotionally intense backchannel responses.

To address these challenges, we propose a cascaded multi-scale individual standardization network that normalizes backchannel expressions relative to individual neutral traits, effectively removing person-specific variations while focusing on generalizable agreement indicators. Besides, we incorporate an augmentation module to alleviate generalization challenges caused by data imbalance. This individual-normalized feature extraction ensures that the model learns agreement patterns that represent relative changes from personal baselines rather than absolute values, significantly improving generalization across diverse expressive behaviors.
Our comprehensive experiments demonstrate that CMIS-Net effectively handles individual differences and data imbalance, achieving state-of-the-art performance. Our analysis also reveals an important distinction between modalities: auditive and visual backchannels. The former tend to carry more intense emotional expressions, present greater challenges for agreement estimation. This finding suggests that future research should move beyond treating backchannels as a unified phenomenon and instead develop modality-specific approaches that account for the unique characteristics of visual and auditive responses.

The main contributions of this paper are summarized as follows:
\begin{itemize}
    \item We propose CMIS-Net, the first cascaded multi-scale individual standardization framework that disentangles individual-specific and invariant features at both frame and sequence levels for backchannel agreement estimation. Beyond backchannel analysis, the proposed framework provides a new approach that can benefit a wide range of subtle pattern recognition tasks.
    \item We design an encoder-decoder-based translator module for sequence-level backchannel features standardization.
    \item We introduce an implicit data augmentation module that addresses distributional bias near agreement score peaks, improving model generalization.
    \item We perform comprehensive ablation studies and visualizations demonstrating the effectiveness of the proposed method, and provide insights into the distinct challenges posed by auditive versus visual backchannels, suggesting they should be treated as separate modalities.
\end{itemize}

\section{Related Work}
\subsection{Backchannel Agreement}
Backchannel was first introduced by Victor Yngve in 1970 \cite{yngve1970getting}. Since then, increasing attention has been paid to backchannel behavior by researchers in linguistics and social psychology \cite{duncan1974structure, bavelas2000listeners}. Early studies on backchannels mainly focused on predicting when a backchannel response occurs \cite{truong2010rule, ruede2019yeah}. However, relatively few studies have addressed the task of identifying whether a behavior sequence contains backchannel signals. The task of Backchannel Agreement Estimation was not formally proposed until 2022, when Philipp Müller et al. \cite{muller2022multimediate} introduced it for the first time. They provided new annotations for the group conversation dataset MPIIGroupInteraction \cite{muller2018detecting}, creating the first public annotations for rating the strength of backchannel agreement. In addition, they established baseline results for this new task. Specifically, they evaluated models using different modalities, including head and facial movements, body posture, and audio features. Their results showed that head motion features achieved the best performance in estimating agreement strength.

Subsequently, Garima Sharma et al. \cite{sharma2022graph} proposed a graph based approach. In their method, group interactions are modeled as a graph, where each participant is a node and edges represent social interactions between individuals. Graph convolution and edge convolution are used to capture both node features and local interaction structures. Two types of graphs were constructed: static graphs, in which each data sample has a single associated graph, and dynamic graphs, in which temporal continuity is modeled by connecting each node to itself across time steps. Similar to Müller et al.'s findings\cite{muller2022multimediate}, head motion features outperformed other modalities in the agreement estimation task.

Moreover, Ahmed Amer et al. \cite{amer2023backchannel} explored the use of different Transformer-based architectures for automatic backchannel detection and agreement estimation \cite{vaswani2017attention}. In their single-stream setting, concatenated body and facial features pass through a Transformer layer. A two-layer stacked version with shared linear prediction achieved the best agreement estimation performance among existing methods.

However, most current works focus primarily on feature selection and multimodal fusion strategies. Although some studies note that predictions may be affected by individual-specific behaviors (e.g., facial expressions), they do not systematically explore the impact of individual differences on backchannel agreement estimation. Therefore, this paper proposes a personalized backchannel agreement estimation framework. Our method disentangles individual-specific neutral features from original representations, allowing the model to focus on features more likely to contain meaningful and generalizable agreement signals to improve prediction robustness.

\subsection{Individual Differences}
Mitigating the influence of individual differences in pattern recognition remains a fundamental challenge. Due to variations in physiology, expressive capability, cultural background, and habitual behaviors, the manner in which different individuals express patterns can differ significantly. Such inter-individual variability often obscures the universal cues that models aim to learn, thereby reducing generalization performance and model robustness \cite{sariyanidi2014automatic, chu2016selective}. Similar challenges are frequently encountered in domains such as micro-expression recognition \cite{hurley2014background}, stress detection\cite{giannakakis2019review} and facial expression recognition \cite{wang2024pose}, where the target signals are transient and highly susceptible to individual-specific traits.

In this context, backchannel agreement recognition provides a representative example of sequential pattern recognition, where individual differences complicate the task. Although individual differences have not yet been explicitly explored in the task of backchannel agreement estimation, existing studies have shown that individuals vary significantly in terms of response frequency, modality preference, emotional intensity, of communication behaviors including backchanneling. 
These differences are influenced by factors such as personality traits, cultural communication norms, and prior conversational experiences \cite{blomsma2024backchannel, de2010influence}. For instance, some individuals may frequently and actively nodding even in low-engagement contexts, while others may withhold agreement cues to enhance persuasive impact. Such individual variation can affect model predictions and potentially reduce generalization.

Previous works attempted to reduce the impact of individual differences by building separate classifiers for different types of individuals \cite{chen2017subject}. However, these methods require maintaining multiple classifiers, leading to increased model complexity, higher training costs, and limited extensibility and generalizability. Some approaches avoid building multiple models by introducing personalized parameters \cite{wang2025multi}
, while others utilize transfer learning \cite{zhang2019individual, zheng2016personalizing} or domain adaptation methods \cite{zhang2022gaussian, wang2021prototype}.
Nevertheless, these approaches typically rely on access to some information from the target domain during training, which is often unrealistic in real-world scenarios. Moreover, their generalization ability may degrade when applied to unseen domains. 

There has been limited discussion on methods that explicitly disentangle individual-specific features and preserve individual-invariant cues during modeling. Evidence from other domains suggests that such disentanglement strategies can enhance model robustness and performance. Previous studies have shown that facial expressions can be modeled as the superposition of a individual-specific neutral face and emotional components \cite{wu2020leed}. Building on similar assumptions, the Individual Standardization Network (ISNet) \cite{fan2022isnet} demonstrated how speech signals can be decomposed into neutral and emotional components. Inspired by these approaches, this paper investigates individual standardization methods for backchannel agreement estimation, focusing specifically on visual features and cascaded multi-scale models. 

\subsection{Cascaded Multi-Scale Modeling}
Cascaded multi-scale architectures refer to models that process information at multiple granularities through cascaded stages. In such designs, lower tiers extract fine-grained features over short intervals, and higher tiers aggregate these into coarser representations. Cascaded multi-scale modeling captures both local and global patterns by integrating hierarchical features across different levels of abstraction. These complementary representations work synergistically to improve the performance.

\cite{roy2021one} propose an MSC-RNN for radar signal classification, with a two-tier RNN architecture that first separates clutter from targets using a short-timescale RNN, and then classifies the target type using a long-timescale RNN. \cite{sordoni2015hierarchical} develop a hierarchical recurrent encoder–decoder for context-aware query suggestion, which encodes sequences of prior queries and generates follow-up queries. This model uses a low-level RNN per query and a high-level RNN for the session, making it sensitive to both immediate and long-range query context. \cite{kuo2018compact} present a compact CNN for facial emotion recognition extended with a frame-to-sequence GRU. 
They first apply a deep CNN to each video frame and then feed the per-frame output distributions into a GRU to model temporal evolution of expressions. This cascaded structure leverages multi-frame context.
In this work, we comprehensively explore the Cascaded Multi-Scale Individual Standardization Network for backchannel agreement estimation.

\section{Cascaded Multi-Scale Individual Standardization Network (CMIS-Net)}

\subsection{Data Pre-Processing}
In our method, we first extract facial landmarks from the input videos using OpenFace 2.0 \cite{tadas2018openface} to minimize background-induced noise. Besides, rather than using raw landmarks directly, we compute inter-frame differences to generate motion-based features $I \in \mathbb{R}^{(M - 1) \times H}$, before feeding them into the CMIS-Net for backchannel agreement detection. This differential representation emphasizes temporal dynamics over static appearance, allowing the model to capture the subtle facial movements characteristic of backchannel responses. This follows the established practices in state-of-the-art backchannel detection research \cite{amer2023backchannel,wang2023unveiling}.

\begin{figure*}[t]
    \centering
    \includegraphics[width=\textwidth]{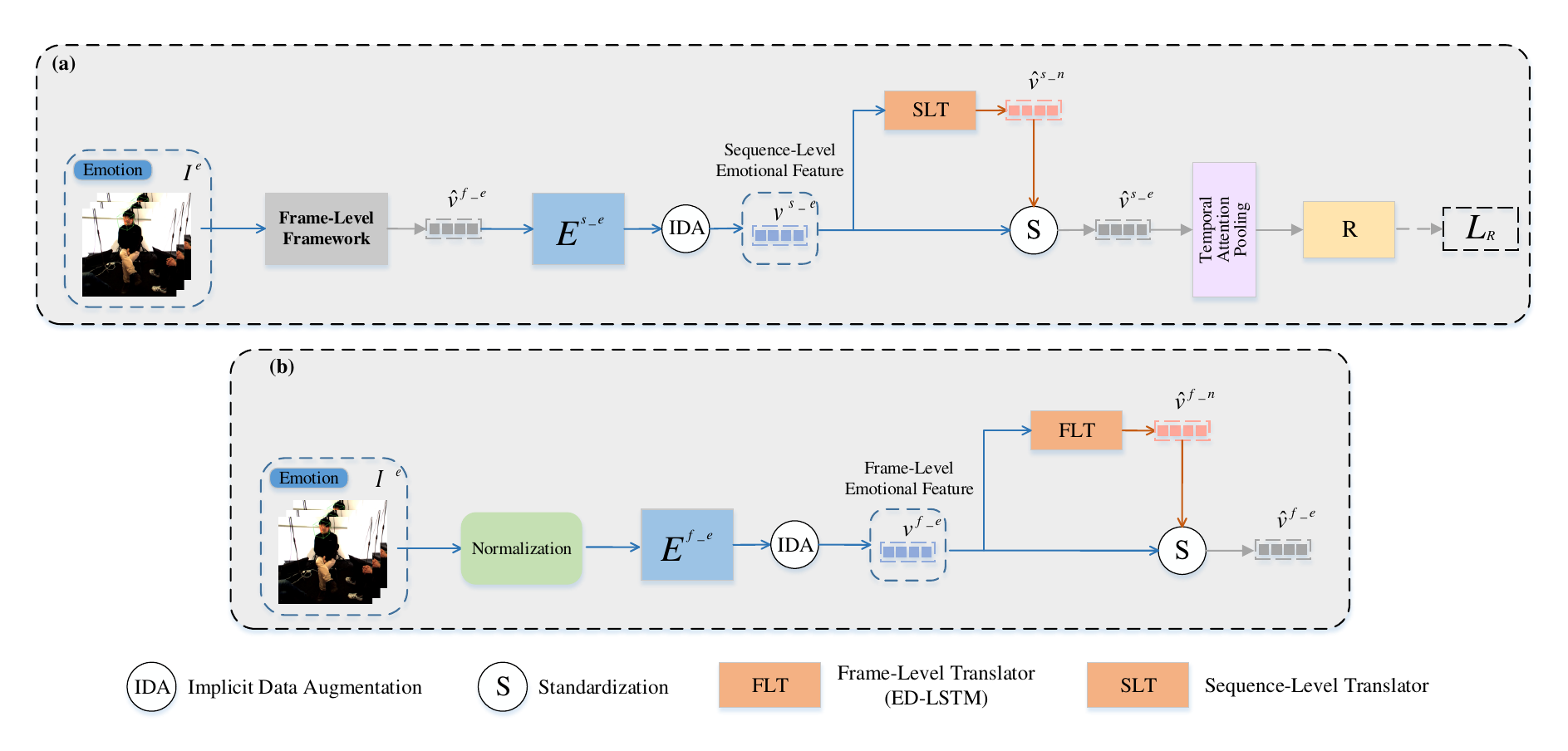}
    \caption{Overview of the CMIS-Net architecture (a), utilizing two cascaded standardization modules: 1. Frame-level standardization (b) extracts instant dynamic features while considering individual-specific variations from each frame; and 2. Sequence-level standardization captures temporal patterns across frames while further disentangling person-specific characteristics. Video inputs are processed through both modules sequentially to obtain generalizable representations for backchannel agreement estimation.}
    \label{fig:framework}
\end{figure*}

\subsection{Architecture and Workflow of CMIS-Net}
Although backchannel behaviors are often regarded as socially normative responses, individual neutral and backchannel expressions can vary across people. For example, some may maintain subtle positive expressions like gentle smiles as their neutral state during conversation, while others naturally exhibit more stoic or impassive facial configurations. These person-specific neutral traits may interfere with model judgment. 
Inspired by previous findings in emotion recognition \cite{fan2022isnet}, we model backchannel agreement expressions as a composite signal containing both individual-specific neutral traits and generalizable backchannel agreement patterns.
Our approach extracts individual-normalized features by removing person-specific neutral baselines from observed expressions. This normalization process eliminates the impacts of individual differences, allowing the model to focus on relative changes from each person's baseline rather than absolute expression variation values.

The proposed multi-scale modeling approach comprises a frame-level individual standardization module that captures instantaneous normalized dynamics and a sequence-level individual standardization module that normalizes temporal patterns. The overall architecture of the proposed model is shown in \hyperref[fig:framework]{Fig.~\ref*{fig:framework}}. The proposed model aims to extract individual-invariant behavioral patterns of backchannel responses from video data. Consistent with prior individual standardization work, we categorize samples based on relative backchannel intensity: ``neutral'' samples contain no backchannel responses, while ``emotional'' samples exhibit relatively higher levels of backchannel behavior. Section \ref{subsec:result_impact_neutral} presents the empirical methodology for this classification.

It is worth noting that the core component of CMIS-Net is the standardization procedure incorporating with translator modules. We systematically evaluated various design configurations and combinations for the components to determine the optimal framework for video-based backchannel agreement prediction.


\subsubsection{The General Translator Framework and Standardization Process}

\begin{figure*}[t]
    \centering
    \includegraphics[width=\linewidth]{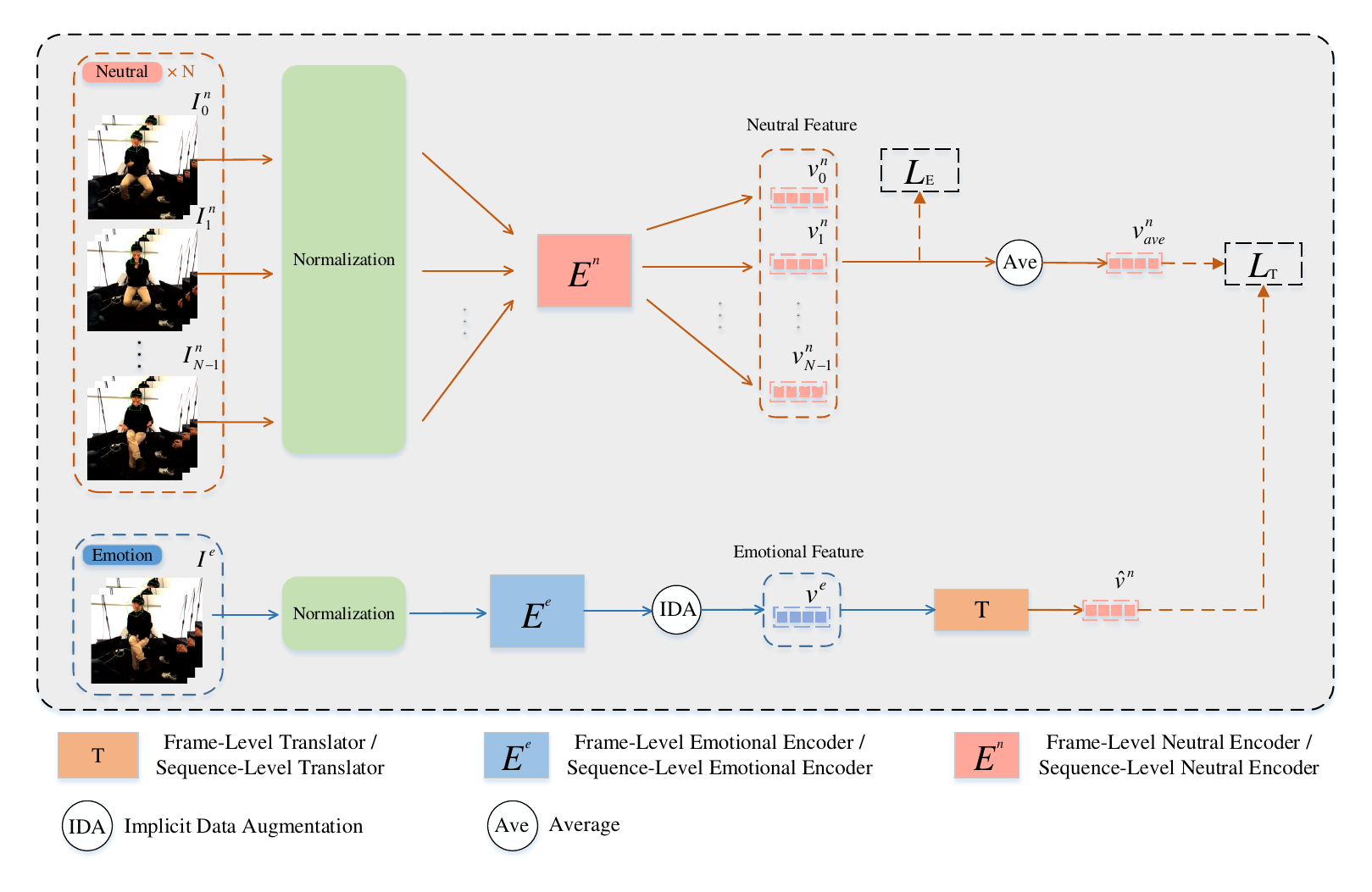}  
    \caption{Training workflow for the translator module and standardization. The neutral feature pipeline (red paths) generates averaged neutral representations from multiple neutral samples, while the emotional feature pipeline (blue paths) performs feature disentanglement through the translator module.}
    \label{fig:trainT}
\end{figure*}

While the frame-level and sequence-level standardization frameworks operate at different scales, they follow a unified training and standardization procedure, as illustrated in Fig. \ref{fig:trainT}. 
Both frameworks employ parallel encoder architectures, $E^n$ and $E^e$ to process neutral (non-backchannel) and emotional (backchannel) samples respectively, enabling consistent feature extraction across temporal scales.

In particular, the neutral encoder ($E^n$) takes $N$ different neutral samples of the same individual as input. Each sample is encoded into a neutral feature vector. By averaging these $N$ vectors, we obtain a statistical benchmark representation that captures individual-specific characteristics.
\begin{equation}
v_i^n = E^n(I_i^n)
\end{equation}
\begin{equation}
v^n_{ave} = \frac{1}{N} \sum_{i=1}^{N} v_i^n
\end{equation}

We denote this as the neutral feature pipeline (marked in red in Fig. \ref{fig:trainT}), which applies to both the Frame-Level Neutral Encoder and Sequence-Level Neutral Encoder modules, i.e., the FLNE ($E^{f\_n}$) and SLNE ($E^{s\_n}$) in Fig. \ref{fig:framework}.

The emotional encoder ($E^e$) takes the original sample as input and encodes it into an emotional feature vector. This vector is then passed to the translator, which aims to disentangle the individual-specific neutral components from the emotional representation.Through the standardizer based on the subtraction operation, the standardized emotional feature vector is generated.
\begin{equation}
v^e = E^e(I_e) 
\end{equation}
\begin{equation}
\hat{v}^e = v^e-T(v^e)
\end{equation}
We refer to this as the emotional feature pipeline (marked in blue in Fig. \ref{fig:trainT}). This pipeline similarly applies to both the frame-level (involving FLEE and FLT) and sequence-level (involving SLEE and SLT) emotional feature encoding.

\subsubsection{The Training Process of Encoders and Translators}

The encoders and translators follow a two-step training process. In the first step, we train the neutral feature pipeline and the emotional encoders within the emotional feature pipeline.
The neutral feature encoder (for both FLNE and SLNE) processes $N$ neutral samples from the same individual as the emotional encoder's input. An approximation loss trains the neutral encoder to extract individual-specific basline expressions:

\begin{equation}
L_E = -\frac{1}{B} \sum_{i=1}^{B} \sum_{j=1}^{N-1} \sum_{j+1}^{N} d(v^n_j,v^n_{j+1})
\end{equation}

where $B$ is the batch size and $d(v^n_j,v^n_{j+1})$ represents the L1 distance between neutral features ${v^n_j}$ and ${v^n_{j+1}}$. This approach enables the neutral encoder to extract robust, stable individual-specific neutral feature representations.

The emotional feature pipeline, operating at both frame and sequence levels. In the first training step, we pre-train emotional encoders at each level (FLEE and SLEE) to directly estimate agreement levels. A temporary regression predictor $\hat{R}$ ensures the encoder learns meaningful features. After training, $\hat{R}$ is removed and the encoder parameters are frozen.
For backchannel agreement prediction, we use a regression loss $L_R$ based on Mean Squared Error (MSE):
\begin{equation}
L_R = -\frac{1}{B} \sum_{i=1}^{B} {(y_i-\hat{y}_i)}^2
\end{equation}
where $y$ represents ground truth labels and $\hat{y}$ denotes predicted scores.

The initial emotional feature training step prepares the emotional encoders for subsequent fine-tuning with translators using $L_T$. The translator loss is designed to extract individual-specific neutral features from emotional representations, preserving information relevant to more generalizable backchannel agreement:
\begin{equation}
L_T = -\frac{1}{B} \sum_{i=1}^{B} d(v^n_{ave},\hat{v}^n)
\end{equation}

In the second training stage (particularly for the emotional feature pipeline), the trained neutral encoder extracts $N$ neutral features from videos of the same individual as $I^e_i$, which are averaged to create a benchmark neutral representation $v^n_{ave}$. The loss measures the distance between this benchmark and the Translator's output neutral features, encouraging the Translator to extract individual-specific neutral features from emotional representations.

This approach enhances the model's ability to improve generalizability across individuals. It encourages the translator to extract individual-specific neutral components from the emotional features. After standardization, it retains the information more likely associated with the relative expression values indicative of backchannel agreement. This disentanglement process allows the model to focus on generalizable communicative cues rather than individual-specific traits, and effectively addressing the challenge of cross-individual generalization.

The neutral feature pipelines are active only during the training phase, where they guide the translator modules (both FLT and SLT) to learn individual-specific baseline features for standardization. During inference, only the emotional feature pipeline operates, performing individualized standardization using the trained translators.

\subsubsection{The Detailed Multi-Scale Process}

The frame-level processing begins with the Frame-Level Neutral Encoder (FLNE, $E^{f\_n}$), which processes $N$ neutral samples ${I_1^n, \ldots, I_{N}^n}$ from the same individual, encoding them into representations $v^{f\_n} = {v_1^{f\_n}, \ldots, v_{N}^{f\_n}}$. FLNE integrates a multi-layer perceptron with frame-wise attention to capture backchannel-related features within individual frames. Operating exclusively during training, this module guides the translator in learning individual-specific neutral features. In parallel, the Frame-Level Emotional Encoder (FLEE, $E^{f\_e}$) processes backchannel samples $I^e$ into emotional representations $v^{f\_e}$, maintaining architectural consistency with FLNE. The Frame-Level Translator (FLT, $T^f$) then extracts neutral representations $\hat{v}^{f\_n}$ from these emotional features through stacked self-attention blocks that operate independently on each frame, refining intra-frame features via gated attention and residual connections. 
A subtraction operation between $v^{f\_e}$ and $\hat{v}^{f\_n}$ extracts relative emotional patterns, yielding generalizable frame-level backchannel representations ($\hat{v}^{f\_e}$) for subsequent modeling.

Building upon frame-level processing, the sequence-level components capture temporal dependencies across frames. The Sequence-Level Neutral Encoder (SLNE, $E^{s\_n}$) encodes frame-level representations into sequence-level representations $v^{s\_n} ={v_1^{s\_n}, \ldots, v_{N}^{s\_n}}$ using a Transformer Encoder backbone to model temporal relationships. Similarly, the Sequence-Level Emotional Encoder (SLEE, $E^{s\_e}$) processes frame-level features to extract emotional temporal patterns ($v^{s\_e} ={v_1^{s\_e}, \ldots, v_{N}^{s\_e}}$) while maintaining structural consistency with SLNE. The Sequence-Level Translator (SLT, $T^s$) employs an encoder-decoder architecture to reconstruct individual-specific neutral feature sequences, $\hat{v}^{s\_n}$, from emotional inputs $v^{s\_e}$, facilitating the disentanglement of individual-specific traits from temporal emotional patterns.

The standardization (i.e., a subtraction operation) is then applied to $v^{s\_e}$ and $\hat{v}^{s\_n}$ to extract relative emotional patterns, yielding generalizable sequence-level backchannel representations ($\hat{v}^{s\_e}$) for subsequent modeling.

\subsubsection{Temporal Attention Pooling Module (TAP) and Regression}
\begin{figure*}[t]
    \centering
    \includegraphics[width=\linewidth]{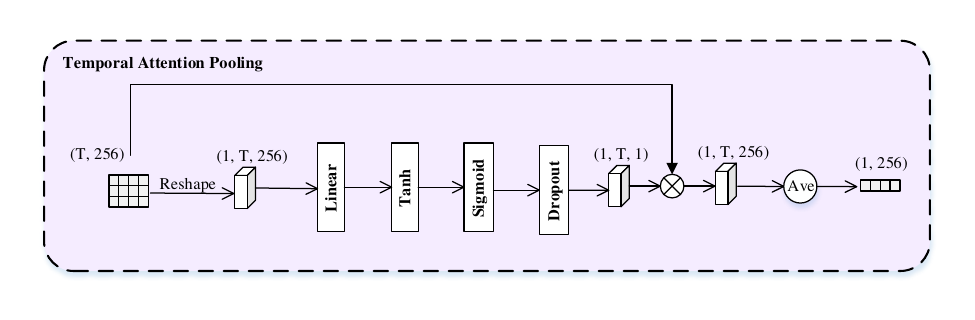}  
    \caption{The Temporal Attention Pooling (TAP) Module.}
    \label{fig:temp_attn}
\end{figure*}
Unlike overt verbal expressions, backchannel are typically brief, occur in specific interactional moments, and are closely tied to the rhythm and timing of the speaker's utterances. These behaviors are not evenly distributed throughout the communication, but appear at specific critical moments. To capture this fleeting and context-dependent behavior, we introduce a temporal attention module. Temporal Attention allows the model to focus on these key moments by calculating weights for each frame. \hyperref[fig:temp_attn]{Fig.~\ref*{fig:temp_attn}} shows the framework of this module

Given a sequence of feature representations after the sequence-level standardization ($\hat{v}^{s\_e} ={\hat{v}_1^{s\_e}, \ldots, \hat{v}_{t}^{s\_e}, \ldots, \hat{v}_{N}^{s\_e}}$), this module applies linear projection to each frame to calculate a scalar attention score and applies a non-linear activation function $\tanh$ to improve expressiveness. A sigmoid is used to generate bounded importance weights.
\begin{equation}
\alpha_t = \sigma(W \cdot \hat{v}_{t}^{s\_e})
\end{equation}

These weights are used to re-weight each frame.
\begin{equation}
\hat{x}_t = \alpha_t \cdot \hat{v}_{t}^{s\_e}
\end{equation}

The final sequence-level representation is obtained by averaging the weighted features over time.
\begin{equation}
z = \frac{1}{T} \sum_{t=1}^{T} \hat{x}_t
\end{equation}


where $T$ is the number of frames. This design enables the network to down-weight uninformative frames while retaining those that contribute most to the target behavior, such as moments where a nod or backchannel utterance occurs.

Finally, a regression model ($R$) takes the feature vector $z$ after the multi-scale standardization and TAP as input to estimate the final agreement level. It is composed of multiple fully connected layers, enabling the extraction of high-level representations for accurate estimation. Similar to $\hat{R}$, $R$ is trained by $L_R$. However, it is worth to note that, only the parameters of $R$ are updated in this step, while the emotional encoder remains fixed.

\subsection{Translator Module}
\subsubsection{Frame-Level Translator (FLT)}

\begin{figure*}[t]
    \centering
    \includegraphics[width=\linewidth]{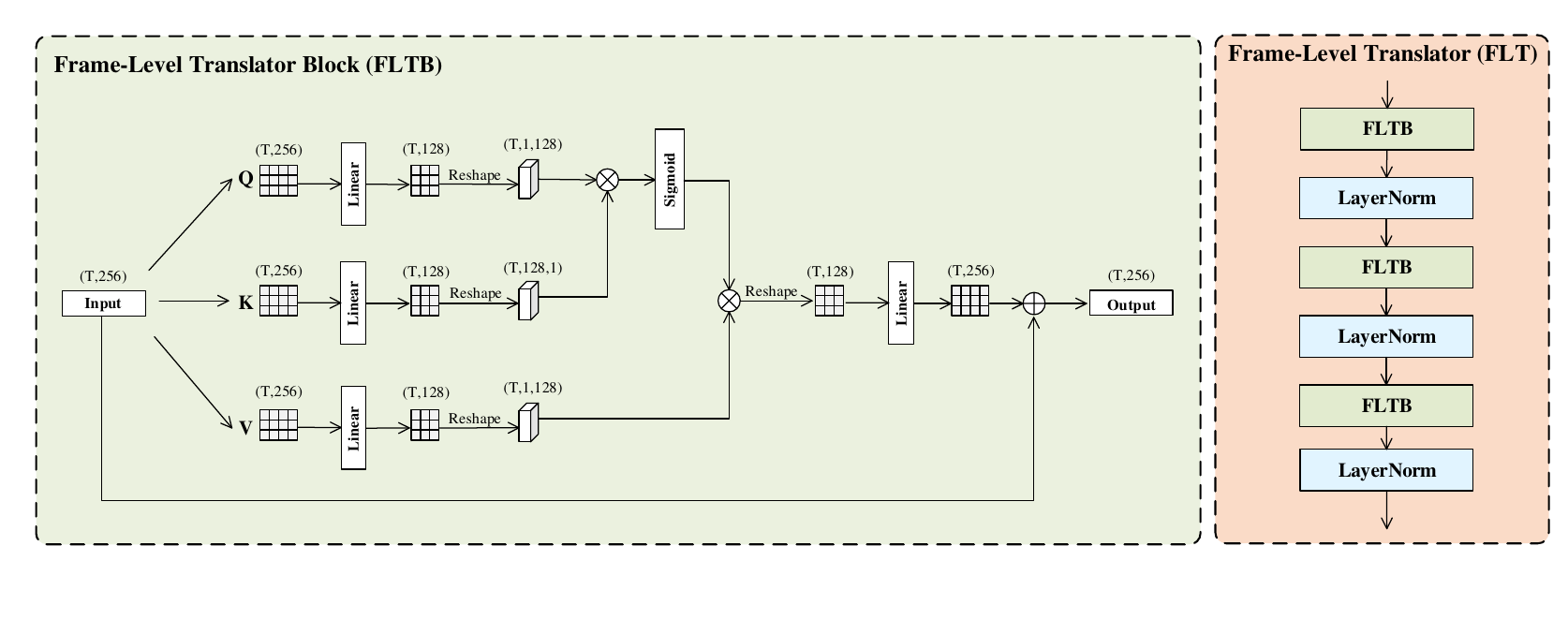}  
    \caption{The FLTB (left) and FLT (right) Module.}
    \label{fig:FLT}
\end{figure*}

Before frame-level encoding, we compute inter-frame differences to capture facial motion. These frame-level features encode individual movements, such as head nods, shakes, lip movements, reflecting the inherently dynamic nature of backchannel behaviors, which are conveyed through subtle motions rather than static expressions.
To extract the individual-specific neutral features in frame-level, we introduce a FLT module, which consists of three stacked Frame-Level Translator Blocks (FLTB) (\hyperref[fig:FLT]{Fig.~\ref*{fig:FLT}}).

FLTB generates query $Q$, key $K$, and value $V$ vectors for each frame via linear projections. Rather than computing cross-frame attention, it calculates a scalar attention score per frame using the dot product between that frame's query and key vectors:
\begin{equation}
\alpha = \sigma(Q, K^T)
\end{equation}
Where, $\sigma(\cdot)$ is the sigmoid function, acting as a soft gate to modulate the value vector:
\begin{equation}
f_{temp} = \alpha \cdot V
\end{equation}
This vector $f_{temp}$ is then linearly projected back to the original input dimension $f_{proj}$ and added to the original frame representation via a residual connection, followed by layer normalization:
\begin{equation}
output = LayerNorm(f_{proj}+input)
\end{equation}


\subsubsection{Sequence-Level Translator (SLT)} \label{sec:slt}
\begin{figure}[h]
    \centering
    \includegraphics[width=0.4\linewidth, angle=90]{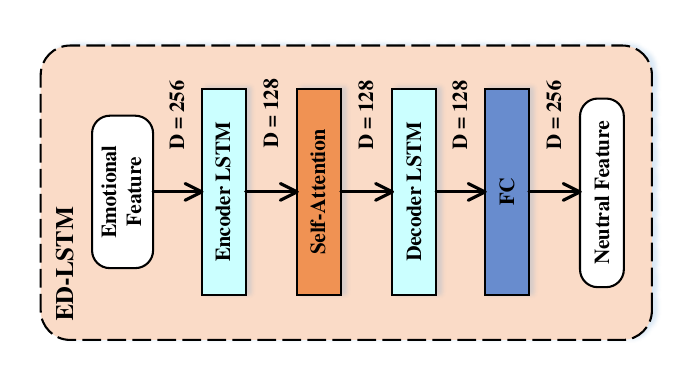}  
    \caption{The ED-LSTM Translator Module.}
    \label{fig:ED-LSTM}
\end{figure}
Although the frame-level standardization module captures useful intra-frame behavioral cues, it fails to account for temporal dynamics across frames. In our framework, the sequence-level features represent the frequency and temporal patterns of these actions over time. Sequence-level patterns—such as nodding frequency, repetition, or rhythm—can provide additional cues that are crucial for estimating backchannel agreement. To capture temporal features, we introduce an Encoder-Decoder LSTM (ED-LSTM) module as SLT, which aims to extract individual-invariant features from the input sequence. As shown in \hyperref[fig:ED-LSTM]{Fig.~\ref*{fig:ED-LSTM}}, the ED-LSTM consists of an encoder LSTM, an attention layer, and a decoder LSTM. 

The encoder module takes a sequence of frame-level features as input and passes them through a unidirectional LSTM to generate a hidden representation. It encodes the underlying sequential structure of the input behavior. To enhance the contextual attention of sequence-level representation, we add an attention module between the encoder and decoder. The decoder receives the attention-weighted context vector as input. This context vector is then passed through an LSTM decoder to reconstruct the sequence-level neutral features from emotional features.



\subsection{Implicit Data Augmentation (IDA)}
During our experiments, we observed that the distribution of backchannel agreement data was imbalanced. Most samples had agreement levels concentrated near the agreement score peak (around 0.25 in the dataset), and the extreme high- and low-level samples were rare. This imbalance led the model to learn local concentrated feature patterns during training, which undermines its ability to generalize to data with different distributions. The detailed analysis is presented in \hyperref[sec:Impact of data augmentation module]{Section~\ref*{sec:Impact of data augmentation module}}.

To tackle this issue, we draw inspiration from Implicit Data Augmentation methods \cite{wang2019implicit,li2021feature,seo2022implicit}, which generate new feature representations in the hidden space instead of directly synthesizing input data. Accordingly, we propose an Implicit Data Augmentation (IDA) module to address this problem. It aims to reduce the model’s reliance on the highly concentrated regions of the data distribution and improve its generalization performance. After the emotional encoder, we introduce this module, as illustrated in \hyperref[fig:framework]{Fig.~\ref*{fig:framework}}. During the training phase, this module adds element-wise Gaussian noise with zero mean and a standard deviation of 0.05 to the output tensor, followed by a random masking operation applied with a probability of 0.1. These operations help model to mitigate overfitting to the training data. 

By adding this module, the model is exposed to more diversities during training, which improves its robustness to edge-case samples, enhancing overall generalizability.


\section{Experiments Details}
\subsection{Dataset}
We use the MPIIGroupInteraction dataset \cite{muller2018detecting, muller2022multimediate} for backchannel agreement estimation. MPIIGroupInteraction dataset is a publicly available dataset that provides annotations of backchannel occurrences and agreements expressed through backchannels in group interactions. The dataset consists of 78 German-speaking participants. Each interaction involves three or four participants engaging in a conversation in a quiet office environment. The dataset includes both audio recordings of the conversations and individual video recordings of each participant. 
In the agreement estimation task, there are 3358 and 1427 annotated video samples in the training and validation sets respectively, each lasts about 10 seconds. Multiple annotators labeled each video, with agreement levels averaged across annotators and normalized to $[-1, 1]$, where $1$ and $-1$ indicate strong agreement and disagreement respectively.

\subsection{Experimental Settings}
Empirical studies from both our work and prior research demonstrate that the relevant backchannel cues typically occur only in the final second of the video samples. Particularly, previous work \cite{amer2023backchannel} use the last three seconds as input yields the most effective results. Shorter intervals may lack sufficient contextual information, whereas longer intervals tend to introduce excessive non-task-related content. Based on this, we use OpenFace 2.0 \cite{tadas2018openface} to extract facial landmarks from each video frame as the initial features. To effectively capture facial dynamics during the critical period, we process only the last 3 seconds of each video and compute the coordinate differences between each frame and its subsequent frame, using the resulting landmark variation rates as input features for the model \cite{wang2023unveiling}. 

Our model is implemented in PyTorch and trained on an NVIDIA GeForce RTX 2080 Ti GPU. The model was trained with the Stochastic Gradient Descent (SGD) optimizer, using its default parameter settings, where the momentum was set to 0.9 and the weight decay coefficient to 1e-4. The batch size was set to 32, and training was performed for a total of 100 epochs. A StepLR learning rate scheduler was employed to decay the learning rate at fixed intervals. The initial learning rate was set to 0.01 and multiplied by 0.1 every 20 epochs. The Transformer encoder consists of 6 stacked Transformer layers, each with 4 attention heads.

\section{Results}
We conduct comprehensive evaluation experiments, which comprises three parts: component contribution analysis, visualization studies on individual standardization effectiveness, and comparative experiments with state-of-the-art methods.
Following the evaluation protocol from \cite{muller2022multimediate}, we train on their training set, evaluate on the validation set, and use Mean Squared Error (MSE) to measure the estimation performance.

\subsection{Ablation Study}
\subsubsection{Impact of the Appropriately Configured SLT}
\label{sec:Translator ablation study}

\begin{table}[ht]
    \centering
    \caption{
    Ablation study of different SLT implementations. ↓ indicates that a lower value corresponds to better model performance.}
    \label{tab:translator experiments}
    \begin{tabularx}{\linewidth}{l >{\centering\arraybackslash}X}
        \toprule
        \textbf{Model} & 
        \makecell[c]{\textbf{Agreement} \\ \textbf{(MSE ↓)}} \\
        \midrule
        Attention-based \protect\cite{fan2022isnet} & 0.058268 \\
        ED-GRU & 0.059717 \\
        ED-LSTM & \textbf{0.057962} \\
        \bottomrule
    \end{tabularx}
\end{table}

In our works, the translator is the key in eliminating individual-specific neutral features. Since human communication behaviors, including both neutral expressions and backchanneling, are context-dependent, temporally sensitive micro-interactions influenced by current and surrounding contextual frames, effective temporal modeling becomes essential for accurately extracting generalizable backchannel agreement features while removing individual-specific neutral patterns.

Previous studies \cite{bahdanau2014neural, sutskever2014sequence} have demonstrated that encoder–decoder architectures possess strong sequence modeling capabilities. Based on this, we adopt such an architecture to construct our translator module. 
We experimented with various temporal models as encoders and decoders to identify the optimal temporal structure for the translator. 
This includes attention-based \cite{fan2022isnet}, GRU-based, and LSTM-based modules (ED-LSTM in Section \ref{sec:slt}, our proposed approach).
The ablation results are presented in \hyperref[tab:translator experiments]{Tabel~\ref*{tab:translator experiments}}. 
 
Our LSTM-based model achieves better performance with an MSE of 0.057962, outperforming all counterparts including the translator model from previous work \cite{fan2022isnet}.
This improvement may be attributed to the ability of ED-LSTM to effectively capture contextual dependencies and generate more representative neutral features. Moreover, in sequence-related tasks, studies have shown that LSTM generally outperforms GRU in capturing temporal dependencies with higher accuracy \cite{britz2017massive}, which aligns with the findings in our task.

\subsubsection{Impact of IDA module}
\label{sec:Impact of data augmentation module}

\begin{figure*}[t]
    \centering
    \begin{subfigure}[b]{0.23\textwidth}
        \centering
        \includegraphics[width=\linewidth]{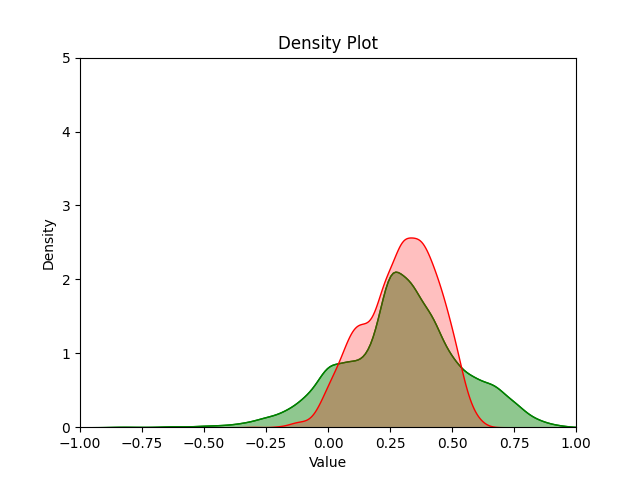}
        \caption*{(a1)}
        \label{fig:a1}
    \end{subfigure}
    \hfill
    \begin{subfigure}[b]{0.23\textwidth}
        \centering
        \includegraphics[width=\linewidth]{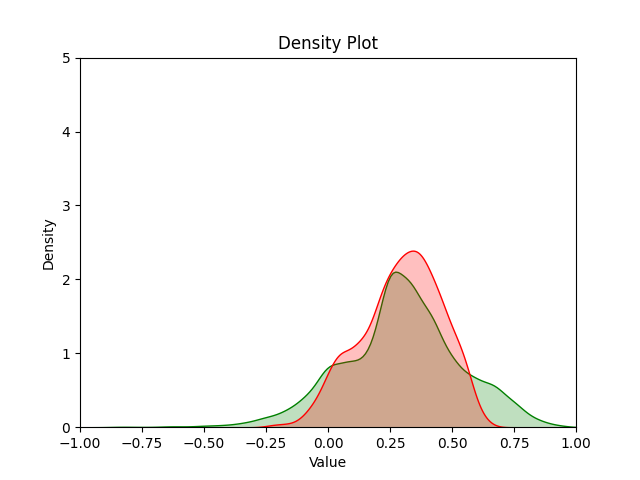}
        \caption*{(a2)}
        \label{fig:a2}
    \end{subfigure}
    \hfill
    \begin{subfigure}[b]{0.23\textwidth}
        \centering
        \includegraphics[width=\linewidth]{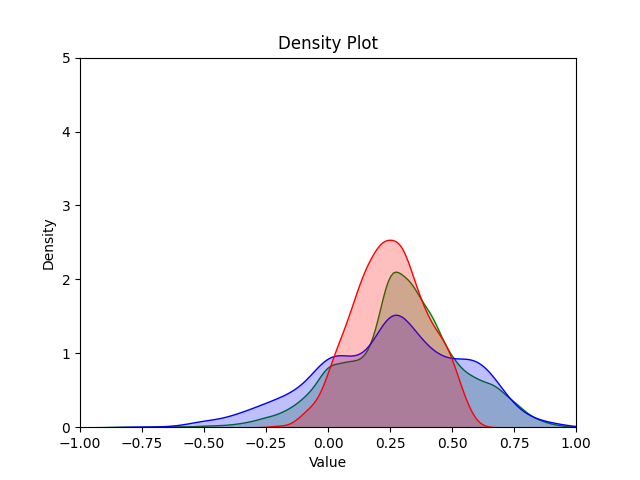}
        \caption*{(b1)}
        \label{fig:b1}
    \end{subfigure}
    \hfill
    \begin{subfigure}[b]{0.23\textwidth}
        \centering
        \includegraphics[width=\linewidth]{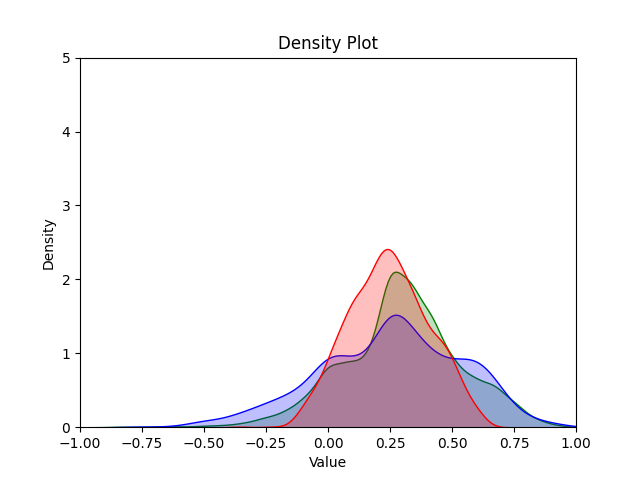}
        \caption*{(b2)}
        \label{fig:b2}
    \end{subfigure}
    \caption{Overall comparison of density distributions before and after applying IDA. Green, blue, and red curves represent the training set, validation set, and predicted results, respectively. In group (a), the red curve shows predictions on the training set; in group (b), on the validation set. Each group includes results before augmentation (a1 and b1) and after augmentation (a2 and b2).}
    \label{fig:4-in-1-horizontal}
\end{figure*}

As shown in \hyperref[fig:4-in-1-horizontal]{Fig.~\ref*{fig:4-in-1-horizontal}} (green curves), the training samples are densely distributed in the range of 0 to 0.5, and few samples show disagreement or strong agreement. In particular, the sample density peaks around 0.25. This uneven distribution causes the model to overly focus on samples near 0.25, resulting in overfitting and reduced generalization ability. Consequently, the prediction results tend to concentrate around 0.25 during inference. As shown in \hyperref[fig:4-in-1-horizontal]{Fig.~\ref*{fig:4-in-1-horizontal}} (b1), this issue becomes even more obvious in the distribution of prediction results on the validation set. The model's prediction range is mainly limited to -0.25 to 0.6, and the density peak around 0.25 is significantly higher than that of the ground truth distribution of the validation set. 

To address this problem, we introduced the IDA module. By adding noise perturbations to the embedding in the latent space, this module helps prevent the model from overfitting. As shown in group (a) and group (b) of \hyperref[fig:4-in-1-horizontal]{Fig.~\ref*{fig:4-in-1-horizontal}}, after applying IDA, the prediction distributions become wider and the peak values decrease. This visualization demonstrates the module's ability to improve the model generalization, which is further confirmed by the results in \hyperref[tab:different augmentation injection strategies]{Tabel~\ref*{tab:different augmentation injection strategies}}, where MSE decreased from 0.059501 to 0.057962 after incorporating the IDA module.

To investigate the effect of IDA, we design four different noise injection strategies. 
\begin{itemize}
    \item \textbf{Non-Augmentation}: Without both FLEE Augmentation (FLEE Aug.) and SLEE Augmentation(SLEE-Aug.). 
    \item \textbf{FLEE Aug. only}: With FLEE Aug. only.
    \item \textbf{SLEE Aug. only}: With SLEE Aug. only.
    \item \textbf{Double Augmentation}: With both FLEE Aug. and SLEE Aug..
\end{itemize}

As shown in \hyperref[tab:different augmentation injection strategies]{Tabel~\ref*{tab:different augmentation injection strategies}}, our results show that applying IDA individually after either the FLEE or the SLEE stage leads to notable performance improvements. Interestingly, the best performance is achieved when noise is injected at both stages simultaneously, indicating that multi-level noise augmentation provides complementary benefits.

\begin{table}[ht]
    \centering
    \caption{Ablation study on different augmentation injection strategies.}
    \label{tab:different augmentation injection strategies}
    \begin{tabularx}{\linewidth}{%
        >{\hspace{6pt}}l<{\hspace{6pt}}     
        >{\centering\arraybackslash}X       
        >{\centering\arraybackslash}X       
    }
        \toprule
        \textbf{FLEE Aug.} & \textbf{SLEE Aug.} & 
        \makecell[c]{\textbf{Agreement} \\ \textbf{(MSE ↓)}} \\
        \midrule
        False & False & 0.059501 \\
        True & False &  0.058641 \\
        False & True &  0.058680 \\
        True & True & \textbf{0.057962} \\
        \bottomrule
    \end{tabularx}
\end{table}

\subsubsection{Impact of different neutral feature selection strategies} \label{subsec:result_impact_neutral}
Another important aspect of our work is the selection of neutral samples. We propose the following two ways of neutral sample selection.

\begin{itemize}
    \item \textbf{Peak-Based Neutral Selection:} We assume that samples with agreement levels concentrated around the peak of the training data distribution (e.g., near 0.25) represent neutral expressions. This assumption is based on the observation in \hyperref[fig:4-in-1-horizontal]{Fig.~\ref*{fig:4-in-1-horizontal}} that the dataset shows a high density of samples in this range, which may indicate a commonly occurring, baseline response state.
    \item \textbf{Non-Backchannel-Based Neutral Selection:} According to the definition provided by (\cite{muller2022multimediate}), positive samples from the backchannel detection task are used as inputs for the agreement estimation task and neutral samples are those without backchannel behavior. Since backchannel agreement inherently involves active feedback behaviors, non-backchannel instances—which lack such signals—can be regarded as neutral samples.
\end{itemize}

In the Peak-Based Neutral Selection, we conducted multiple experiments using different ranges centered around 0.25. For example, if the edge is set to 0.1, the selected range for neutral samples becomes 0.15 to 0.35. As shown in  \hyperref[tab:different neutral]{Tabel~\ref*{tab:different neutral}}, the best performance was achieved when edge equals 0.1. The results of the Peak-Based Neutral Selection slightly outperformed those of the Non-Backchannel-Based Neutral Selection. Since both selection methods demonstrated promising results, we consider these two neutral selection strategies to be reasonable and effective.

\begin{table}[ht]
    \centering
    \caption{Ablation study of different neutral feature selection strategies. The actual selection range for neutral samples is defined as neutral ± edge. The second-best results are indicated with an underline.}
    \label{tab:different neutral}
    \begin{tabularx}{\linewidth}{%
        >{\hspace{6pt}}l<{\hspace{6pt}}     
        >{\centering\arraybackslash}X       
        >{\centering\arraybackslash}X       
    }
        \toprule
        \textbf{Neutral} & \textbf{Edge} & 
        \makecell[c]{\textbf{Agreement} \\ \textbf{(MSE ↓)}} \\
        \midrule
        0.25 & 0 & 0.057967 \\
        0.25 & 0.1 & \textbf{0.057915}  \\
        0.25 & 0.25 & 0.058005 \\
        non-backchannel & - & \underline{0.057962} \\
        \bottomrule
    \end{tabularx}
\end{table}

\subsubsection{The contribution of each component}

\begin{table*}[ht!]
    \centering
    \caption{
    Ablation study of different module combinations.}
    \label{tab:contribution of each component}
    \hspace*{-0.04\textwidth}
    \begin{tabularx}{1.08\textwidth}{
    >{\centering\arraybackslash}X 
    >{\hspace{6pt}}l<{\hspace{6pt}} 
    >{\centering\arraybackslash}X 
    >{\centering\arraybackslash}X}
        \toprule
        \textbf{Scale} &
        \textbf{Model} & 
        \textbf{Pooling Method} & 
        \makecell[c]{\textbf{Agreement} \\ \textbf{(MSE ↓)}} \\
        \midrule
        \multirow{8}{*}{Single-Scale}
            & \multirow{2}{*}{$FLE+R$} 
                & Global Pooling  & 0.061117\\
                & & TAP & \textbf{0.060800} \\
            \cmidrule(lr){2-4}
            & \multirow{2}{*}{$FLE+FLT+R$}
                & Global Pooling & 0.061852 \\
                & & TAP & \textbf{0.061346} \\
            \cmidrule(lr){2-4}
            & \multirow{2}{*}{$SLE+R$}
                & Global Pooling & \textbf{0.059432} \\
                & & TAP & 0.060807 \\
            \cmidrule(lr){2-4}
            & \multirow{2}{*}{$SLE+SLT+R$}
                & Global Pooling & \textbf{0.059249} \\
                & & TAP & 0.061777 \\
        \hline
        \multirow{8}{*}{Multi-Scale}
        & \multirow{2}{*}{$FLE+SLE+R$} 
                & Global Pooling  & 0.058259 \\
                & & TAP & \textbf{0.058246} \\
        \cmidrule(lr){2-4}
        & \multirow{2}{*}{$FLE+FLT+SLE+R$} 
                & Global Pooling  & 0.058543 \\
                & & TAP & \textbf{0.058069} \\
        \cmidrule(lr){2-4}
        & \multirow{2}{*}{$FLE+SLE+SLT+R$} 
                & Global Pooling  & \textbf{0.058083} \\
                & & TAP & 0.058393 \\
        \cmidrule(lr){2-4}
        & \multirow{2}{*}{$FLE+FLT+SLE+SLT+R$} 
                & Global Pooling  & 0.058165 \\
                & & TAP & \textbf{0.057962} \\
        \bottomrule
    \end{tabularx}
\end{table*}

We conduct ablation studies to investigate the contribution of each module in our framework, including the impact of single- and multi-scale individual standardizations, as shown in Table~\ref{tab:contribution of each component}.
In the single-scale setup, the best performance was achieved by the sequence-level module, reaching an MSE of 0.059249. It suggests that a frame-level module alone may struggle to capture sufficient backchannel-related information. Given that backchannel behaviors are typically expressed in a continuous and temporally dependent manner, this observation aligns with intuitive expectations.

Meanwhile, multi-scale configurations consistently outperformed single-scale ones, indicating that combining frame-level and sequence-level features provides complementary information.
Although each single-scale module may have limited capacity to capture all relevant cues, their combination enhances the model’s ability to estimate backchannel agreement more accurately, achieving the best MAE of 0.057962. 

Moreover, the results show that performance is also influenced by the choice of pooling method.
Specifically, in the single-scale setting, adding SLT to the sequence-level module resulted in degraded performance under TAP. However, this should not be interpreted as SLT failing to extract meaningful neutral features. Interestingly, when using global average pooling, the same SLT-enhanced module achieved better performance. This aligns with prior research suggesting that temporal pooling doesn't universally guarantee better results, as global pooling can sometimes more effectively integrate information across entire sequences.

\subsection{Visualizing the Effect of CMIS-Net}
\begin{figure*}[t]
    \centering
    \begin{subfigure}[b]{0.23\textwidth}
        \centering
        \includegraphics[width=\linewidth]{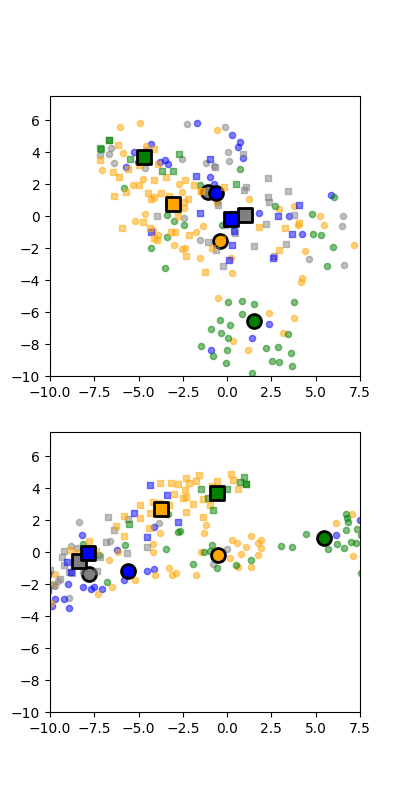}
        \caption*{(Group 1)}
        \label{fig:rec27_pos2vsrec10_pos4}
    \end{subfigure}
    \hfill
    \begin{subfigure}[b]{0.23\textwidth}
        \centering
        \includegraphics[width=\linewidth]{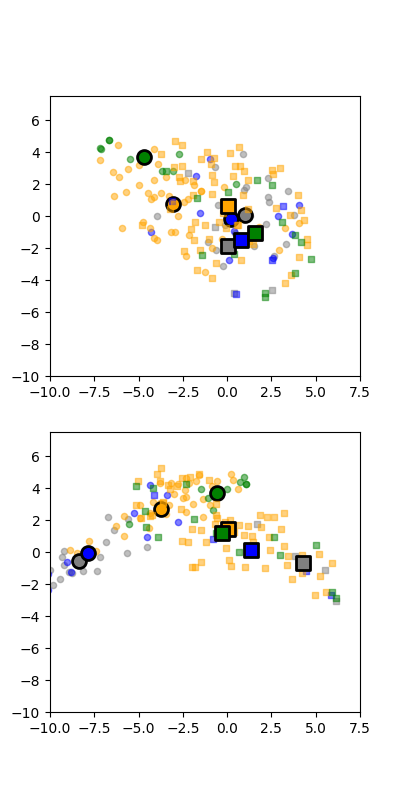}
        \caption*{(Group 2)}
        \label{fig:rec10_pos4vsrec26_pos3}
    \end{subfigure}
    \hfill
    \begin{subfigure}[b]{0.23\textwidth}
        \centering
        \includegraphics[width=\linewidth]{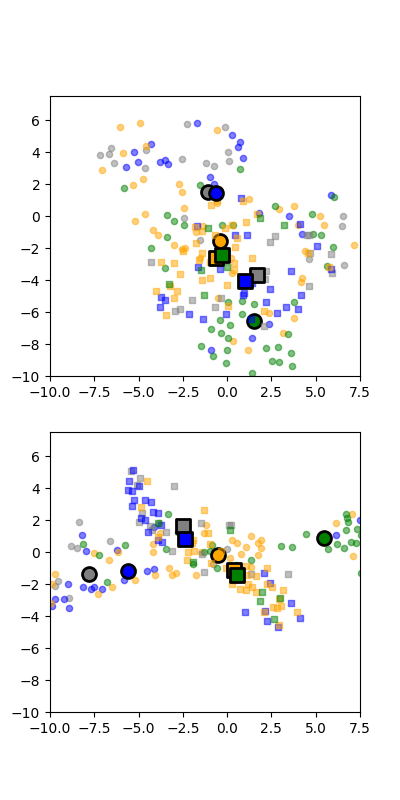}
        \caption*{(Group 3)}
        \label{fig:rec27_pos2vsrec09_pos2}
    \end{subfigure}
    \hfill
    \begin{subfigure}[b]{0.23\textwidth}
        \centering
        \includegraphics[width=\linewidth]{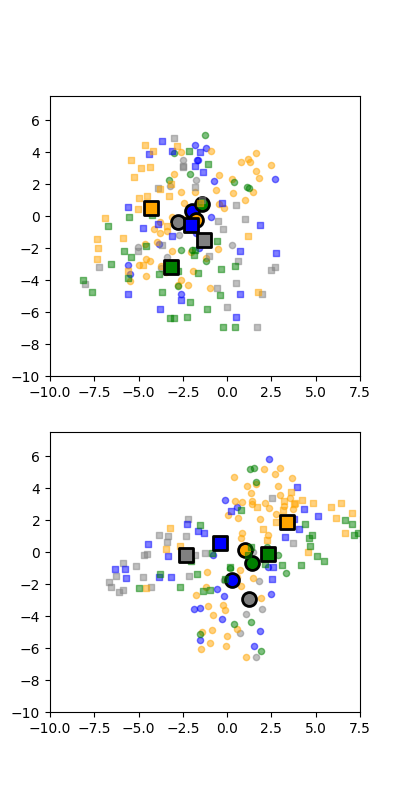}
        \caption*{(Group 4)}
        \label{fig:rec10_pos1vsrec28_pos2}
    \end{subfigure}
    \caption{Feature distribution visualization across four groups before (top) and after (bottom) standardization. Circles and squares represent two individuals in each group. The continuous agreement scores are divided into four ranges: [-1, 0) (gray), [0, 0.25) (blue), [0.25, 0.5) (orange), and [0.5, 1] (green). The highlighted markers indicate the center points of each ranges.}
    \label{fig:visual_IS}
\end{figure*}

To intuitively demonstrate the effectiveness of our proposed framework and the individual standardization process, we conducted a feature visualization experiment. Specifically, we applied t-distributed Stochastic Neighbor Embedding (t-SNE) to project the high-dimensional features into a 2D space.

As shown in \hyperref[fig:visual_IS]{Fig.~\ref*{fig:visual_IS}}, the top and bottom rows present the features before and after standardization respectively.
For better visualization, we divide the agreement levels into four categories, each represented by a different color. Considering the imbalanced distribution of agreement levels shown in \hyperref[fig:4-in-1-horizontal]{Fig.~\ref*{fig:4-in-1-horizontal}}, we discretized the continuous agreement levels into four categories:
[-1, 0) (gray), [0, 0.25) (blue), [0.25, 0.5) (orange), and [0.5, 1] (green). Each color corresponds to a specific category, and each shape (circle/square) denotes a different individual, with highlighted centroids.

From \hyperref[fig:visual_IS]{Fig.~\ref*{fig:visual_IS}}, we observe that the standardization process brings centroids of the same agreement category closer across different individuals. 
For example, in Group 1, the green and gray centroids from different individuals exhibit greater alignment after standardization. Similar patterns are observed in Group 2 (green), Group 3 (blue and gray) and Group 4 (green). 
The scatters show that before standardization, samples from different individuals and agreement categories are distributed chaotically. After standardization, same-category samples cluster together (blue and orange samples form distinct clusters in Group 1, so as do orange and gray in Group 2, and blue and orange in Group 3). This demonstrates that standardization significantly reduces emotional confusion between individuals.

\begin{table*}[t]
    \centering
    
    \caption{Minimum inter-individual distances between different agreement category centers. 'Cir' and 'Squ' denote the two individuals. 'Total' sums all eight centers. 'Diff' shows the change after standardization.}
    \label{tab:closest distance to each center}
    {\footnotesize 
    \setlength{\tabcolsep}{3pt} 
    \renewcommand{\arraystretch}{1.2}
    \hspace*{-0.25\textwidth}
    \begin{tabularx}{1.5\textwidth}{c c *{10}{>{\centering\arraybackslash}X}}
    \toprule
    Session & Standardizer & cir-gray & cir-blue & cir-orange & cir-green & squ-gray & squ-blue & squ-orange & squ-green & Total & Diff \\
    \midrule
    
    \multirow{2}{*}{1} & Before & \textbf{2.1471} & 2.1658 & 1.5262 & \textbf{6.4991} & 2.1658 & \textbf{1.5262} & 2.1554 & \textbf{4.2834} & 22.4691 & 7.6024\\
    & After & 1.3562 & \textbf{2.8574} & \textbf{3.8946} & 4.2780 & \textbf{2.8574} & 1.3562 & \textbf{4.2780}  & 3.8945 & 29.8962  & \\ 
    \midrule
    
    \multirow{2}{*}{2} & Before & 1.2266 & 0.8562 & \textbf{4.0563} & \textbf{5.6219} & 1.6771 & 1.6722 & 0.8562 & 1.4975 & 17.4640 & 21.8426 \\
    & After & \textbf{8.3268} & \textbf{7.7027} & 3.7777 & 2.3715 & \textbf{6.5848} & \textbf{4.0793} & \textbf{2.3715} & \textbf{3.7777} & 38.9919 & \\ 
    \midrule
    
    \multirow{2}{*}{3} & Before & 4.0838 & 3.9391 & 0.8974 & 2.6147 & \textbf{2.9577} & \textbf{2.6147} & 4.1103 & 0.8974 & 22.1150 & 7.3894 \\
    & After & \textbf{5.8803} & \textbf{4.1884} & \textbf{1.6558} & \textbf{5.5097} & 2.7116 & 2.1230 & \textbf{5.5097} & \textbf{1.6558} & 29.2342 & \\ 
    \midrule
    
    \multirow{2}{*}{4} & Before & 0.7924 & 1.9781 & 0.4674 & 1.5524 & 1.4585 & 0.4674 & 1.8737 & \textbf{2.9527} & 11.5428 & 7.3963 \\
    & After & \textbf{3.1452} & \textbf{2.6557} & \textbf{1.3655} & \textbf{2.2644} & \textbf{3.0486} & \textbf{1.5073} & \textbf{3.2729} & 1.3655 & 18.6252 & \\ 
    \bottomrule
    \end{tabularx}
    }
\end{table*}

While standardization effectively reduces individual bias by eliminating person-specific neutral patterns, complete separation of emotional states remains challenging, particularly in 2D spaces as shown by partial overlap between orange circles and green squares in Group3. This aligns with prior findings~\cite{fan2022isnet}, though such ambiguity in 2D visualization may be distinguishable in the original high-dimensional space.

We calculate the minimum inter-individual distance between each agreement category center of one individual and all different agreement category centers, except the target category, of another individual. This measures how well different agreement categories are separated across individuals in the feature space. 
In the visualization (Fig. \ref{fig:visual_IS}), two shapes (circles and squares) represent two different individuals, while four colors indicate different agreement categories. Taking the gray circle center as an example, we compute distances from this center to all non-gray square centers and select the minimum value. 
Larger minimum distances indicate better separation between agreement categories across individuals, suggesting features can more effectively distinguish agreement levels.
In general, the minimum distances increase after standardization (Table~\ref*{tab:closest distance to each center}), demonstrating improved category separation.

\subsection{State-of-the-Art Comparison}
\begin{table}[ht]
    \centering
    \caption{The performance of different methods.}
    \label{tab: SOTA comparison}
    \begin{tabularx}{\linewidth}{%
        >{\hspace{6pt}}l<{\hspace{6pt}}     
        >{\centering\arraybackslash}X       
    }
        \toprule 
        \textbf{Authors} & \makecell[c]{\textbf{Agreement} \\ \textbf{(MSE ↓)}}\\ \midrule 
         Garima Sharma et al. \cite{sharma2022graph} & 0.066 \\
         Ahmed Amer et al. \cite{amer2023backchannel} & 0.0644 \\
         Baseline 2022: Head Pose \cite{muller2022multimediate} & 0.075 \\
         Baseline 2022: All Features \cite{muller2022multimediate} & 0.079 \\
         Baseline 2022: Trivial \cite{muller2022multimediate} & 0.085 \\
         \midrule 
        \textbf{Ours} & \textbf{0.057962} \\
        \bottomrule 
    \end{tabularx}
\end{table}

We evaluated CMIS-Net against recent methods on the benchmark backchannel agreement validation set. As shown in Table~\ref*{tab: SOTA comparison}, our method achieves the best MSE of 0.057962 -- the first to drop below 0.060 for this task. This improvement are mainly attributed to addressing individual differences and data distribution issues that existing methods overlook while focusing on multi-modal fusion. Notably, our approach achieves superior performance even with a single-scale standardization.

\section{Discussion and Future Work}
Backchannel communication includes visual and auditory modalities \cite{muller2022multimediate}. Particularly, people typically express polite agreement during interactions, and often with auditory feedback (e.g., "yes," "right," "no") conveying stronger agreement responses. This explains the label distribution peak around 0.25, with fewer samples showing disagreement or strong agreement. In our study, we piloted with Light-ASD \cite{liao2023light} model to separately analyze the impact of visual and auditory backchannels to agreement estimation.
It generates frame-level scores (negative for silence, positive for speech) at 25 fps. Following \cite{muller2022multimediate}'s criterion that auditory backchannel events last $\geq 0.25$ seconds, we classified samples as auditory if seven consecutive frames in the last three seconds had scores $>0.2$; otherwise, they were classified as visual backchannel samples.

As shown in \hyperref[tab:Statistics]{Tabel~\ref*{tab:Statistics}}, we obtained 2792 visual backchannel samples and 566 auditive backchannel samples, totaling 3,358 instances in the training set. We then computed the absolute values of the agreement scores for all samples and performed statistical analysis. The results indicate that auditive samples show higher average emotion intensity than visual samples. This trend is consistently observed across the 25th, 50th, and 75th percentiles. These findings suggest that auditive backchannel are more likely to convey stronger emotional intensity compared to visual backchannel. In order to gain deeper insights, we conducted the following experiments:

\begin{itemize}
    \item \textbf{all-to-all:} All available samples are used for both training and validation.
    \item \textbf{all-to-visual:} The training set includes all (visual + auditive) samples, while the validation set consists only of visual backchannel samples.
    \item \textbf{visual-to-all:} The training set consists only of visual backchannel samples, while the validation set includes all samples.
    \item \textbf{visual-to-visual:} Both the training and validation sets consist exclusively of visual backchannel samples.
\end{itemize}

\begin{table}[ht]
    \centering
    \caption{Statistics of auditive backchannel labels in the training set.}
    \label{tab:Statistics}
    \begin{tabularx}{\linewidth}{>{\hspace{6pt}}l<{\hspace{6pt}} >{\centering\arraybackslash}X >{\centering\arraybackslash}X}
        \toprule
        \textbf{} & 
        \textbf{Visual} & 
        \textbf{Auditive} \\
        \midrule
        \textbf{Count}   & 2792        & 566 \\
        \textbf{Mean}    & 0.297910    & \textbf{0.442274} \\
        \textbf{Std}     & 0.182404    & 0.225720 \\
        \textbf{Min}     & 0.000000    & 0.000000 \\
        \textbf{25\%}    & 0.167000    & 0.250000 \\
        \textbf{50\%}    & 0.250000    & 0.417000 \\
        \textbf{75\%}    & 0.417000    & 0.667000 \\
        \textbf{Max}     & 0.917000    & 0.917000 \\
        \bottomrule
    \end{tabularx}
\end{table}

\begin{table}[ht]
    \centering   
    \caption{Performance comparison when training on visual backchannel samples: evaluated on visual-only vs. all (visual + auditive) samples.}
    \label{tab:visual backchannel performance}
    \begin{tabularx}{\linewidth}{>{\hspace{6pt}}l<{\hspace{6pt}} >{\centering\arraybackslash}X >{\centering\arraybackslash}X}
        \toprule
        \textbf{Mode} & 
        \makecell[c]{\textbf{Agreement} \\ \textbf{(MSE ↓)}} \\
        \midrule
        \textbf{all-to all} & 0.057962 \\
        \textbf{visual-to-all} & 0.058857 \\
        \textbf{all-to-visual} & \textbf{0.053065} \\
        \textbf{visual-to-visual} & 0.053844 \\
        \bottomrule
    \end{tabularx}
\end{table}

In the backchannel agreement estimation experiments described above, the all-to-visual setting yields the lowest MSE. As shown in  \hyperref[tab:visual backchannel performance]{Tabel~\ref*{tab:visual backchannel performance}}, the performance of the all-to-visual setting surpasses that of the all-to-all setting. It indicates that the model has already achieved relatively good predictive accuracy on visual backchannel samples, whereas its performance on auditory samples remains suboptimal. 
This may be due to the limited auditory backchannel samples, which constrains the model’s ability to generalize to such cases.
As previously discussed, auditive backchannels are often associated with stronger emotional expressions, making accurate prediction of these high-intensity samples especially meaningful.  Moreover, the observation that the visual-to-all setting performs slightly worse than the all-to-all setting suggests that there may be a distributional gap between visual and auditory backchannel samples. Therefore, it is necessary to explicitly consider both visual and auditory backchannels in protocol modeling.

\section{Conclusions}
In this paper, we propose a novel cascaded multi-scale individual standardization framework for backchannel agreement estimation that relieves individual differences and data imbalance. By disentangling individual-normalized backchannel agreement features features from observed cues. 
Our model reduces the interference of personal neutral response patterns and achieves more consistent agreement prediction across diverse participants. 
Additionally, we introduce an implicit data augmentation strategy in the latent space to mitigate sample distribution bias and improve the model’s generalization, particularly in low-frequency, emotionally intense cases (strong dis/agreement). 

Our experiments confirm that individual standardization significantly improves backchannel agreement estimation. Ablation studies and visualizations further validate the contribution of each proposed module. Furthermore, with the comprehensive analysis we identify auditive backchannel samples as inherently more challenging due to their association with stronger affective signals. This finding underscores the necessity of considering visual and auditive backchannels separately in the future. Beyond this specific task, our work has broader implications for pattern recognition by multi-scale individual standardization, potentially benefiting various sequential recognition tasks.

\section{Acknowledgments}
This work was supported, in part, by the Education University of Hong Kong under Grant RG 73/2024-2025R, the Hong Kong Polytechnic University under Grant P0048656, and the Hong Kong Research Grant Council under Grant 15600219.

\bibliographystyle{elsarticle-num} 
\bibliography{cas-refs} 

\end{document}